# BrAIcht, a theatrical agent that speaks like Bertolt Brecht's characters

Roland Baz
LAMSADE, Paris Dauphine - PSL

Kristina Malyseva
ZHDK, Zurich Univ. of the Arts

Anna Pappa
LIASD Univ. Paris 8

Tristan Cazenave
LAMSADE, Paris Dauphine - PSL.

## Abstract

This project introduces BrAIcht, an AI conversational agent that creates dialogues in the distinctive style of the famous German playwright Bertolt Brecht. BrAIcht is fine-tuned using German LeoLM, a large language model with 7 billion parameters and a modified version of the base Llama2 suitable for German language tasks. For fine-tuning, 29 plays of Bertolt Brecht and 907 of other German plays that are stylistically similar to Bertolt Brecht are used to form a more diverse dataset. Due to the limited memory capacity, a parameter-efficient fine-tuning technique called QLoRA is implemented to train the large language model. The results, based on BLEU score and perplexity, show very promising performance of BrAIcht in generating dialogues in the style of Bertolt Brecht.

## 1  Introduction

In the rapidly evolving landscape of technology and creativity, chatbots and conversational agents have changed the way we interact with art (Miller, 2023), music (Hennecke, 2022), dance (Berenguer, 2016) and theater (Dorsen, 2023). They have evolved from simple tools into true artistic collaborators. With their advanced language capabilities and contextual awareness, these conversational agents foster creative innovation by engaging in meaningful dialogues and contributing to artistic endeavors. In theater, conversational agents act as dynamic performers, leading dialogues and interacting with audiences during live performances, while musicians collaborate with chatbot composers to produce unique music. In the visual arts, conversational agents assist artists in brainstorming and curating exhibitions.

This study examines whether AI can reproduce Brecht's unique style in conversational agents and poses the question: Can AI help create conversational agents that reflect Brecht's style? Motivated by the idea of using artificial intelligence in plays, this research aims to create a conversational agent that enables the bot to communicate in the style of dialogues from Brecht's plays.

## 2 Background

Following the success of the transformer archi- tecture by (Vaswani et al., 2017), the AI commu- nity has focused on developing open- and closed- source large language models (LLMs) that com- municate effectively with users. (Radford et al., 2019) introduced GPT-2, a 1.5B parameter model trained on a large collection of web text, which excels at multiple NLP tasks without task-specific fine-tuning, including text completion and ma- chine translation. (Touvron et al., 2023) later de- veloped Llama2, a collection of LLMs with 7B to 70B parameters, optimized for dialogues, show- ing superior performance over other open-source models. Additionally, (Jiang et al., 2023) in- troduced Mistral-7B, which outperforms Llama2 (13B) and Llama1 (34B) in reasoning, math, and code generation tasks, offering faster inference and efficient sequence processing.

Pre-trained AI models can be fine-tuned for spe- cific tasks, improving performance. For example, (Yang, Tang, & Tam, 2023) developed InvestLM, based on Llama-65B, for financial topics like SEC filings and quantitative finance. It performs well in answering investment questions, matching ad- vanced models like GPT-4. Similarly, (Shoham & Rappoport, 2023) created CPLLM for clinical dis- ease prediction, outperforming models like Med- BERT in metrics such as PR-AUC and ROC-AUC. (Yu et al., 2023) used LLMs to tackle challenges in financial time series analysis, showing superior re- sults over traditional models like ARMA-GARCH and gradient boosting.

In the field of theater and the arts, there are few projects that use LLM fine-tuning to generate po- ems and plays in the style of playwrights. One example is (Bangura, Barabashova, Karnysheva, Semczuk, & Wang, 2023), which presents a GPT-2 model for automatically generating German drama texts. First, scene outlines are generated based on keywords, and then these outlines are converted into complete scenes. They use the German the- ater corpus and the German text archive. While the quantitative results are promising, the quality of the generated texts is rated as poor, likely due to the quality of the training data.

A notable contribution is MoliAIre (Cazenave, Grosjean, Rozie`re, & Pappa, 2022), a conver- sational agent that mimics the dialogue style of Molie`re's characters. The model fine-tunes GPT-2 on Molie`re's corpus to generate responses in the 17th-century French theatrical style. Key im- provements include integrating works from other playwrights for better stylistic coherence and us- ing reverse-generation for improved rhyme gener- ation. Based on previous work, our research offers a significant and unique contribution to the field.

## 3 Analysis and Results

Following the state-of-the-art, BrAIcht uses a generative language model that has been pre- trained on an extensive German text corpus so that it has basic knowledge of German grammar and vocabulary. The parameters of the model are then fine-tuned using a specific corpus consisting of Brecht's works and plays by other artists that have a similar style to Bertolt Brecht. This corpus con- sists of dialog excerpts formatted as follows:

- USER : line 1
- BrAIcht : line 2
- USER : line 3
- BrAIcht : line 4
.

Once trained, the model can function as a conver- sational agent as follows:

- The user enters a prompt. A context is cre- ated that is identical to the training context, but BrAIcht's prompt is left blank.

- This context is fed into the model, which then generates the response from BrAIcht and stops when a specific end-of-prompt to- ken is encountered.

- This process is repeated to generate a con- versation by accumulating the exchanges be- tween the user and BrAIcht.

### 3.1 Data Pre-processing and Encoding

The dataset includes 29 plays by the famous German playwright Bertolt Brecht and 907 plays by other German playwrights who have a similar style to Bertolt Brecht. From this dataset, we extract the cues. We then create the prompts that are provided to the model. We also introduce certain special tokens to better convey the structure of the data. First, we insert a sentence start token la- bel <s> to signal the beginning of the sentence. Similarly, an end-of-sentence token labeled </s> is inserted to signal the end of the sentence. We also insert a pad token labeled <pad>. Finally, the keywords are converted into tokens and used as input to the model.

Our approach involves a two-stage fine-tuning process. In the first training session, we fine-tune the LLM using the German theatre dataset. We then perform a further fine-tuning of the resulting model using the Brecht dataset. We opt for this two-step approach because the Brecht dataset is relatively small. To make the dataset more diverse, we include the German plays that have stylistic similarities with Brecht's works. For the first fine- tuning step, we split the data into a training set (80%) and a validation set (20%). In the second fine-tuning step, however, we use a split of 90% for training and 10% for validation. More detailed information about the split of the data set can be found in Table 1.

| Number of cues | Training set | Validation set |
|---|---|---|
| German plays | 542,474 | 433,979 | 108,495 |
| Brecht plays | 17,740 | 15,966 | 1,774 |

Table 1: BrAIcht datasets

### 3.2 Pre-trained model

We first try the base models of Llama2 and Mistral, but we only get average results. This is because the dataset is in German and Llama2 and Mistral are primarily pre-trained on an En- glish corpus. To solve this problem, we found a

German-focused language model called LeoLM, which builds on Llama2. LeoLM is available in two versions, 7B and 13B, both trained with a context length of 8k. It uses techniques such as linear RoPE scaling and Flash Attention 2 to im- prove training efficiency. To improve the model in German, a second stage of pre- training is per- formed with Llama2 weights and a German cor- pus with 65B tokens. This approach significantly improves the performance of the model in Ger- man compared to the Llama2 and Mistral baseline models.

### 3.3 Parameter-efficient Fine-tuning

The model is loaded and trained on an A6000
GPU with 48 GB VRAM. However, since fine- tuning all the parameters of the model is an ex- tremely resource-intensive process that requires considerable computing power, we decide to fine- tune the smallest LeoLM model with 7B pa- rameters. While the larger models are more promising in their results, they require signifi- cantly more memory and computing power that our single GPU cannot provide. In addition, we use parameter-efficient fine-tuning (PEFT), known as Quantized Low-rank Adaptation (QLoRA), to train the model. This approach allows us to op- timize the performance of the model and adapt it to our task while reducing the high memory re- quirements and training time associated with fine- tuning. The main idea behind the QLoRA tech- nique developed by (Dettmers, Pagnoni, Holtz- man, & Zettlemoyer, 2023) is that it uses a novel high-precision technique to quantize a pre-trained model to 4-bit (Frantar, Ashkboos, Hoefler, & Al- istarh, 2023), (Frantar & Alistarh, 2022), and then adds a small set of learnable low-rank adapter weights that are tuned by backpropagating gradi- ents through the quantized weights. Consequently, this technique reduces the size of the model so that it fits into the GPU while maintaining performance comparable to full fine-tuning.

### 3.4 Training

To train the base LeoLM-7B, we first obtain its weights from Hugging Face, a widely used plat- form where the AI community shares their open- source models and contributes to the development and progress of AI research (Wolf et al., 2020).

#### 3.4.1 Objective Function and Optimization

The learning objective is to reconstruct the di- alog extracts by minimizing the logarithm of per- plexity, which is a common metric used to evaluate language models, where lower values indicate bet- ter performances in predicting the next word in a sequence. Each dialog extract is divided into a se- quence of tokens $U = (u_1, \ldots, u_n)$ and the model parameters $\Theta$ are optimized by minimizing the in- verse of the log-likelihood function (negative log-likelihood):

$$L(U, \Theta) = -\sum_{i=1}^{n} \log P(u_i \mid u_1, \ldots, u_{i-1}, \Theta) \quad (1)$$

To minimize the objective loss function $L(U, \Theta)$ from equation 1, we use an optimization tech- nique called AdamW (Zhuang, Liu, Cutkosky, & Orabona, 2022), (Zhou, Xie, & YAN, 2023) a variant of Adam, an optimization technique com- monly used for training machine learning and deep learning models.

### 3.4.2 Inference

To promote the diversity of the model's re- sponses and improve its improvisational ability, we use a stochastic generation strategy. Specif- ically, we use the top-k sampling method for text generation in combination with temperature, which controls the randomness of predictions. Based on the context prompt, the model generates a probability distribution over all possible next to- kens. We then select the next token from the 50 most probable options (k = 50 and temperature = 0.7).

### 3.4.3 Evaluation

The results show a significant reduction in per- plexity, reaching a value of 3.57 on the validation dataset after the second stage of fine-tuning. The results are shown in Table 2.

| | German plays | Brecht plays |
|---|---|---|
| Perplexity | 10.19 | 3.57 |

Table 2: Perplexity

We proceed with the calculation of the BLEU score, keeping in mind that this metric can be in- fluenced by various parameters such as top-p, top- k, temperature, and the number of generated can- didates (n). While BLEU is traditionally used for evaluating translation tasks, it has limitations in evaluating dialogue models, particularly because it focuses on n-gram overlap rather than creative or stylistic coherence. Nevertheless, it provides a useful point of comparison for assessing improve- ments in model performance and allows for a stan- dardized evaluation metric across different mod- els. To simplify our approach, we generate only one candidate per reference and set top-k to 50. In our case, we compare the model tuned to German plays with the model tuned to Brecht's plays. This comparison is performed using the Brecht valida- tion set to assess how well each model generates Brecht's plays. We take samples of size 100, 300, 500, and 1000 from the dataset and calculate the BLEU score three times for each sample, taking the average of the results. The final scores are shown in Table 3. To further illustrate this, we perform three BLEU score trials for each sample size and then average the three results. We repeat this process for all sample sizes. At the end, we calculate the average of all the BLEU scores obtained to get the final score, which clearly shows a significant improvement from 0.26 to 0.665.

| | German plays | Brecht plays | |
|---|---|---|---|
| | 0.46 | 0.70 | n=100 |
| | 0.22 | 0.65 | n=300 |
| | 0.18 | 0.57 | n=500 |
| | 0.17 | 0.74 | n=1000 |
| BLEU (avg) | 0.26 | 0.665 | |

Table 3: BLEU score

## 4 Conclusion & Recommendations

The aim of this research project is to develop a conversational agent that is able to generate plays in the style of Bertolt Brecht using large language models. For this purpose, the LeoLM-7B, a German variant of the Llama2, was selected due to its strong performance on German language tasks. The fine-tuning process consisted of two phases: First, the model was trained on a dataset of German plays and then retrained exclusively on plays by Brecht. This training was carried out with an A6000 GPU and 48 GB RAM. The preliminary results are promising and show that the model can generate plays that reflect Brecht's unique style. The model achieved a BLEU score of 0.67 and a perplexity of 3.57, indicating solid performance. However, a major limitation in this project was the availability of GPU resources, which affected the model selection. Future work could improve the accuracy of the model through fine-tuning larger models and the use of retrieval-augmented generation techniques.